\documentclass[journal]{IEEEtran}

\usepackage{times}  
\usepackage{helvet} 
\usepackage{courier}  
\usepackage[hyphens]{url}  
\usepackage{graphicx} 
\usepackage{caption} 
\usepackage{amsmath}
\usepackage{amssymb}
\usepackage{algorithmic}
\usepackage[ruled]{algorithm2e}
\usepackage[pagebackref=true,breaklinks=true,letterpaper=true,colorlinks,bookmarks=false]{hyperref}
\usepackage{makecell}
\usepackage{multirow}

\begin{document}

\title{Joint Multi-Dimension Pruning via Numerical Gradient Update}

\author{Zechun Liu,
        Xiangyu Zhang,
        Zhiqiang Shen$^{\dagger}$,
        Yichen Wei,
        Kwang-Ting Cheng,~\IEEEmembership{Fellow,~IEEE,}
        and Jian Sun
        
\thanks{Zechun Liu is with Hong Kong University of Science and Technology, Hong Kong, China. e-mail: zliubq@connect.ust.hk This work was done when Zechun Liu interns at MEGVII Technology.}
\thanks{Xiangyu Zhang, Yichen Wei, Jian Sun are with MEGVII Technology, Beijing, China. email: \{zhangxiangyu, weiyichen, sunjian\}@megvii.com}
\thanks{Zhiqiang Shen is the corresponding author and is with Carnegie Mellon University, Pittsburgh, PA, USA. email: zhiqians@andrew.cmu.edu}
\thanks{Kwang-Ting Cheng is with Hong Kong University of Science and Technology, Hong Kong, China. e-mail: timcheng@ust.hk.}}

\maketitle

\begin{abstract}
We present joint multi-dimension pruning (abbreviated as JointPruning), an effective method of pruning a network on three crucial aspects: spatial, depth and channel simultaneously. To tackle these three naturally different dimensions, we proposed a general framework by defining pruning as seeking the best pruning vector (i.e., the numerical value of layer-wise channel number, spacial size, depth) and construct a unique mapping from the pruning vector to the pruned network structures. Then we optimize the pruning vector with gradient update and model joint pruning as a numerical gradient optimization process. To overcome the challenge that there is no explicit function between the loss and the pruning vectors, we proposed self-adapted stochastic gradient estimation to construct a gradient path through network loss to pruning vectors and enable efficient gradient update. We show that the joint strategy discovers a better status than previous studies that focused on individual dimensions solely, as our method is optimized collaboratively across the three dimensions in a single end-to-end training and it is more efficient than the previous exhaustive methods. 
Extensive experiments on large-scale ImageNet dataset across a variety of network architectures MobileNet V1\&V2\&V3 and ResNet demonstrate the effectiveness of our proposed method. For instance, we achieve significant margins of 2.5\% and 2.6\% improvement over the state-of-the-art approach on the already compact MobileNet V1\&V2 under an extremely large compression ratio.
\end{abstract}

\begin{figure*}[t]
\centering
\includegraphics[width=0.85\linewidth]{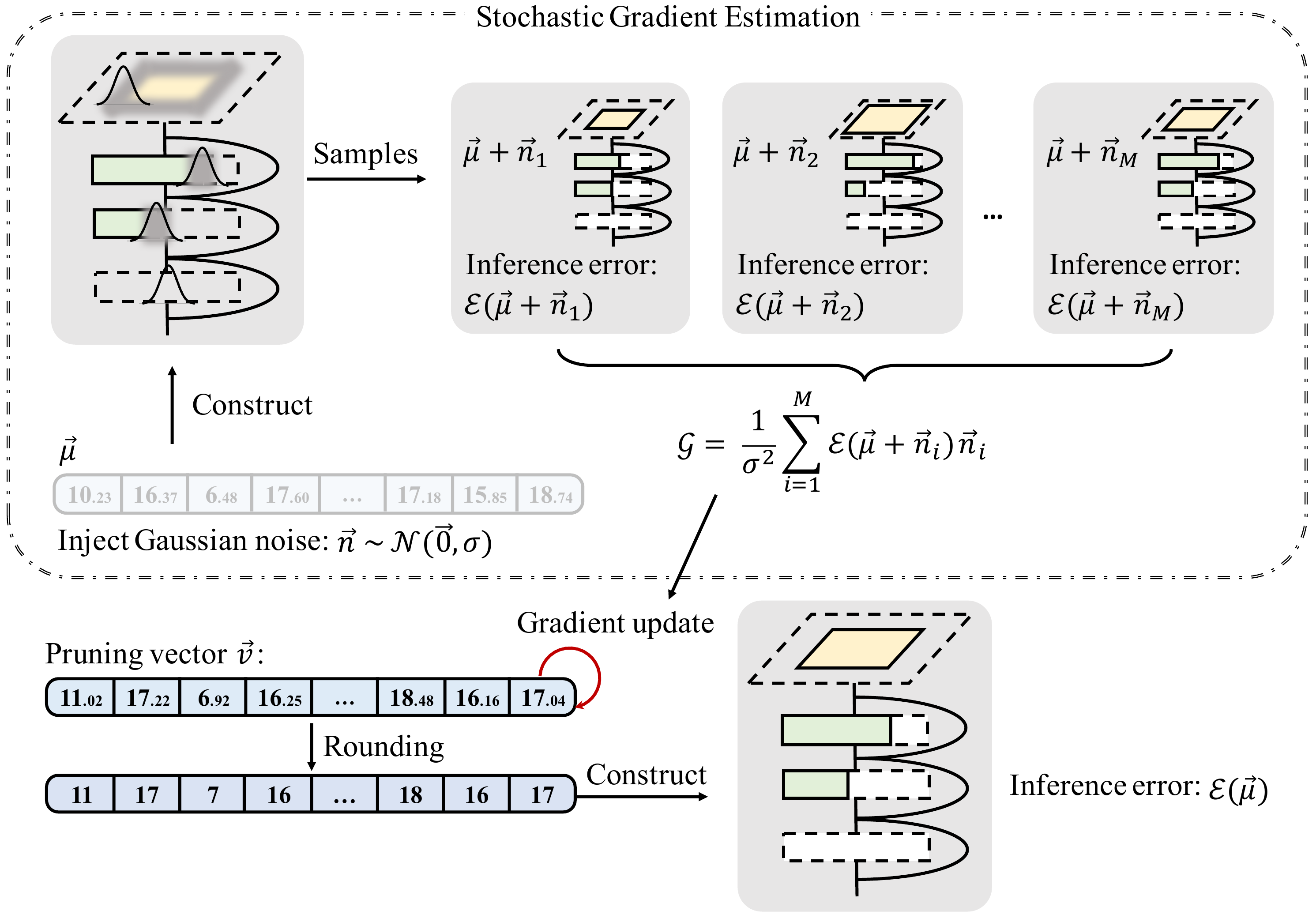}
\caption{The overall framework of the proposed joint multi-dimensional pruning. We formulate the joint pruning to be finding the optimal numerical value of the pruning vector specifying layer-wise channel numbers, spatial resolution and depth. We construct the pruned networks corresponding to the pruning vectors. Then, we randomly sample architectures following Gaussian distribution and calculate the error rates of network architectures in the adjacent loss landscape of the current pruning vector. We use the estimated gradients to numerically update the pruning vectors.}
\label{fig:figure_overall}
\end{figure*}

\section{Introduction}
\label{sec:intro}
Network pruning has been acknowledged as one of the most effective model compression methods for adapting heavy models to the resource-limited devices~\cite{he2017channelpruning,slimmable,networkslimming}. The pruning methods evolved from unstructured weight pruning~\cite{guo2016dynamic, han2015learning} to structured channel pruning~\cite{ding2019centripetal,zhuang2018discrimination} with the purpose of more hardware-friendly implementation. 

With the increasing demand for highly compressed models, merely pruning the channel dimension becomes insufficient to strike a good computation-accuracy trade-off. 
Some previous studies begin to seek compression schemes besides the channel dimension. 
For example, OctaveConv~\cite{chen2019drop} proposed to manually reduce the spatial redundancy in feature maps. MobileNet~\cite{howard2017mobilenets} scales down the input image’s spatial size for reducing the computational overhead. Besides the spatial dimension, depth is also a dimension worth exploring~\cite{crowley2018closer}. These three dimensions (i.e., channel, spatial, depth) are interrelated: A reduction in the network depth can enable using larger feature maps or more channels under the same computational constraint; An alteration in the spatial size will affect the optimal channel pruning scheme. To this end, manually set the spatial size and network depth while only adjust the layer-wise channel pruning strategy leads to sub-optimal solutions. It is of high significance to automatically allocate computational resource among three dimensions with a unified consideration. 

However, jointly pruning channels together with the spatial and depth dimension is challenging and some difficulties are always underestimated. The first challenge is that most of the pruning algorithms are developed based on the unique property of the channel dimension, for instance, removing channels with small weight L1 norms~\cite{networkslimming,yang2018netadapt}. Such approaches are inflexible and can hardly generalize to other dimensions, especially the spatial dimension. The second challenge stems from the overladen optimization space within the three integrated dimensions, since the potential choices of these dimensions are numerically consecutive integers. Therefore, the previous enumerate-based methods are always computationally prohibitive or under-optimized. Such as the EfficientNet~\cite{tan2019efficientnet} adopts the grid search, a typical enumerate-based approach, to find the compound scaling factor of channel, spatial, and depth. This algorithm involves intensive computational overhead to find only three numerical ratios (\textit{i.e.,} depth/width/resolution) and becomes unpractical to generalize to more fine-grained space, \textit{e.g.} the layer-wise pruning ratio, combining with depth and resolution.

To resolve the aforementioned limitations and derive a decent solution, we propose the JointPruning that considers three dimensions (channel, spatial, depth) simultaneously. For the channel dimension, in contrast to the traditional pruning methods that mainly focus on exploring the correlation between channels (\textit{e.g.,} the redundancy), we target at discovering the optimal pruning ratio in each layer, following~\cite{rethinkpruning,liu2019metapruning}, instead of identifying pruning which channel and we directly seek the optimal pruned network structure, \textit{i.e.,} optimal channel numbers in each layer. Then we formulate the joint multi-dimension problem by defining channel numbers associating with the spatial and depth into an integrated space through a pruning vector, specifying the optimal numerical values of channel numbers, spatial resolution and depth. Within this joint space, we can not only mine the relationships inside a single dimension, but also find relatively optimal options across three dimensions by optimizing this pruning vector.
That is to say, we consider the global reciprocity between the three aspects and model joint pruning with the defined pruning vector specifying layer-wise channel numbers, spatial size, and the network depth. This circumvents the algorithmic dependency on the exclusive properties in the channel dimension.

How to find the optimal pruning vector is non-trivial, as the potential combinations are astronomical. This complexity makes simply using the grid search or random sampling infeasible. In this work, we proposed a novel pruning vector update mechanism, which directly optimizes the numerical values of the pruning vector with gradient update. As there is no explicit function between the network loss function of the pruned network and its structural pruning vectors ({\em i.e.}, the network loss function is a function dependent on network weight parameters and the input images, but not the function directly associated with the vector defining the network structure.) To overcome this challenge, we propose self-adapted stochastic gradient estimation to approximate the gradient \textit{w.r.t.} pruning vectors. 
As shown in Fig.~\ref{fig:figure_overall}, we define a pruning vector to specifying the network channel numbers, depth and resolution, then we can have a unique mapping from the pruning vector to the corresponding pruned networks. Then, in stochastic gradient estimation, we randomly inject the Gaussian noise to the pruning vector and evaluate the error of the sampled structures. We use the error rate and the injected noise to approximate the optimal direction of updating the pruning vector. Note that, the pruning vector evaluation process is built atop our proposed weight sharing mechanism, which enables evaluating the pruning vectors without training each corresponding pruned network from scratch. Details for the weight sharing are elaborated in Sec.~\ref{sec:method}.

We show the effectiveness of JointPruning on MobileNet V1/V2 and ResNet structures. It achieves up to $9.1\%/6.3\%$ higher accuracy than MobileNet V1/V2 baselines when targeting three dimensions together under FLOPs constraint. Moreover, JointPruning is friendly to tackle channels in shortcuts and also supports multiple resource constraints. It can automatically adapt the pruning scheme according to the underlying hardware specialty and different circumstances due to its larger potential options. Under CPU latency constraint, JointPruning achieves competitive accuracy with much less optimization time than advanced adaptation-based ChamNet~\cite{dai2018chamnet}. When targeting at GPU latency, it surpasses MobileNet V1/V2 baselines with more than 3.9\% and 2.4\% improvement under an extremely large compression ratio.

Contributions:
\begin{itemize}
    \item We propose a new joint-pruning framework that tackle pruning three different dimensions by modeling the pruning as finding the optimal architecture vector specifying the numerical number of the input resolution, layer-wise channel number, and depth. 
    \item We propose gradient estimation to approximate the gradient with respect to the non-differentiable architecture vectors for update, which can efficiently deal with multiple dimensions at the same time.
    \item Our approach is demonstrated to be effective, flexible, and greatly speed up the searching process as well as occupy much smaller memory footprint, compared with previous methods of the similar direction.
\end{itemize}

\section{Related Work}
\label{sec:related_work}
Model compression is recognized as an effective approach for efficient deep learning~\cite{han2015deep,wu2019efficient}. The categories of model compression expands from pruning~\cite{he2017channelpruning, huang2018data}, quantization~\cite{zhuang2019effective,zhuang2019training}, to compact network architecture design~\cite{ma2018shufflenet,zhang2018shufflenet,howard2017mobilenets,sandler2018mobilenetv2}. Our approach is most related to pruning.

\textbf{Traditional Pruning} 
Early works~\cite{lecun1990optimal,hassibi1993optimal,guo2016dynamic} prune individual redundant weights and recent studies focus on removing the entire kernel~\cite{networkslimming, luo2017thinet} to produce structured pruned networks.
Traditional pruning can mostly be categorized into progressive pruning~\cite{he2017channelpruning, luo2017thinet, networktrimming, he2018soft} and mask-based pruning~\cite{networkslimming, huang2018data,ding2019centripetal,luo2018autopruner,zhuang2018discrimination}. Progressive pruning identifies pruning which channels in a layer-by-layer manner, based on certain metrics, such as the L1-norm of the weight matrix. However, such metrics can hardly be generalized to spatial dimension and these pruning methods involve human participation in deciding the pruning ratio for each layer. 
The mask-based pruning models pruning as selecting useful channels and update the mask with gradients~\cite{networkslimming, he2018amc, huang2018data}. For example, Liu et al.~\cite{networkslimming} proposed to regularize the scaling parameters in BatchNorm and prune the channels associated with small scaling factors. He et al.~\cite{he2018amc} proposed to use a DDPG policy to decide the pruning ratio in each layer. Huang et al.~\cite{huang2018data} further extend the pruning space from channel to discover the layer-wise and block-wise sparsity. However, these mask-based pruning again can hardly be applied to pruning the spatial dimension.

\textbf{Model Adaptation} While traditional pruning involves massive human participation in determining the pruning ratio, AutoML-based methods with reinforcement learning~\cite{chen2019storage, he2018amc}, a feedback loop~\cite{yang2018netadapt} or a meta network~\cite{liu2019metapruning} can automatically decide the best pruning ratio.
In those pruning methods, the optimization space is confined to channels only. Intuitively, a higher compression ratio and better trade-off can be obtained through jointly reducing channel, spatial and depth dimensions.
So far, only a few works address the joint adaptation across different dimensions. A very recent work CC~\cite{li2021towards} proposed to joint channel pruning and tensor decomposition with global compression rate optimization and achieved superior performance gains over previous works focusing on single dimensions. EfficientNet~\cite{tan2019efficientnet} focuses on scaling all dimensions of depth/width/resolution with a single compound coefficient. The coefficient is obtained by grid search. X3D~\cite{feichtenhofer2020x3d} proposed a step-wise network expansion approach for efficient video recognition, which uses coordinate descent to expand a single axis in each step. ChamNet~\cite{dai2018chamnet} uses the Gaussian process to predict the accuracy of compression configurations in three dimensions. Despite their high accuracy, grid search or Gaussian process requires training hundreds of networks from scratch, which are with high computational cost.
Instead, we propose to model the joint pruning as optimizing the numerical values of architecture configurations (i.e., the number of channels, depth and spatial size), which enables to use a \textit{stochastic gradient estimation} \textit{w.r.t.} the pruning vector in optimization, and further improves the efficiency greatly.

\textbf{Gradient Estimation}
Gradient estimation is initially proposed for optimizing non-differentiable objectives~\cite{nesterov2017random}. Reinforcement learning (RL) algorithms often utilize gradient estimation to update the policy~\cite{williams1992simple,o2016combining, gu2017interpolated,lillicrap2015continuous,beyer2002evolution,salimans2017evolution,back1991survey}. We customize a relaxation of the objective function and use the {\em log likelihood trick} to calculate the estimated gradient~\cite{silver2014deterministic}.
Different from these algorithms designed for the typical RL tasks, such as Atari and MuJoCo, our approach focuses on optimizing the network configuration and assumes the policy function to be a Gaussian distribution, which can better approximate the actual gradient and is free of hyper-parameters. Furthermore, we utilize the Lipchitz continuous property of the neural networks \textit{w.r.t.} the pruning vectors. Such that we adopt a progressively shrinking Gaussian window for obtaining more accurate gradient estimations. In this way, we novelly customize the gradient estimation to the network multi-dimension pruning tasks, and further we use an alternative paradigm for updating pruning vectors and the weight parameters. 

\textbf{Neural Architecture Search}
Different from pruning tasks, which aim to find a compressed model through adjusting the numerical values in architecture configurations, NAS are targeting at choosing the best options~\cite{zoph2018learning,real2018regularized,tan2018mnasnet,xie2018snas,bender2018understanding}  and/or connections~\cite{xie2019exploring}. Although a few NAS studies include two or three channel number choices in their search space~\cite{wu2018fbnet,cai2018proxylessnas}, the way of modeling different channels as independent candidates restrict the search space from being extended to consecutive integer channel or spatial choices. In order to achieve consecutive channel number search, MobileNet V3~\cite{howard2019searching} adopts neural architecture search followed with NetAdapt channel pruning~\cite{yang2018netadapt} to explore the efficient structure in a coarse-to-fine manner. 
The gradient-based neural architecture search (NAS) models architectural choices as independent operations~\cite{cai2018proxylessnas, liu2018darts}, while for the pruning task, each channel number or spatial size or depth choice is essentially correlated numerical choices, \textit{i.e.}, if the 20-channel choice is known to perform well, then 19-channel and 21-channel are also likely to perform well. Thus modeling these correlated architecture choices as independent, as the previous gradient-based NAS works, fails to encode the correlation. Instead, the proposed JointPruning directly encodes and updates the numerical values of channels / spatial size / depth. This allows the training algorithm to start with the current architecture configuration and update the numerical value based on how each dimension contributes to the final accuracy improvement. Gradient estimation is proposed to update the configuration vectors. 

\textbf{Parameter Sharing} Different from early NAS algorithms~\cite{cai2018proxylessnas, wu2018fbnet} that store the channel choices as separate options and update the mask associated with it, we design a specific weight-sharing mechanism supporting the direct gradient update to the numerical values of the architecture configuration. Inspired by the recent practice in NAS~\cite{guo2019single,stamoulis2019single,cai2019once}, we adopt matrix-level weight sharing, which only stores one matrix for each layer that contains the maximum channel number choice and crops the weights according to the selected channel number during the training. Incorporated with spatial rescale and drop depth, this weight-sharing mechanism can flexibly accommodate the search in three different dimensions. More details are described in the Methodology.

\setlength{\tabcolsep}{1pt}
\begin{table}[h]
\begin{center}
\caption{Notations}
\label{table:notation}
\resizebox{.45\textwidth}{!}{
\begin{tabular}{ccccccccc}
\noalign{\smallskip}
\hline
Notation & Definition\\
\hline
$c$ & Number of channels in each layer\\
$s$ & Input spatial resolution of the network\\
$d$ & Network depth \\
$w$ & Weight parameters\\
$\vec{v}$ & Pruning vector $\vec{v} = \{ c, s, d\}$\\
$\mathcal{L}(w;\vec{v})$ & Loss function of the neural network\\
$\mathcal{C}(\vec{v})$ & Computational cost of the neural network\\
$\mathcal{E}(w;\vec{v})$ & Error function of the neural network\\
$\mathcal{E}(\vec{v})$ & Abbreviation for $\mathcal{E}(w;\vec{v})$\\
$p_{\theta}(\vec{v})$ & Probability distribution of $\vec{v}$ \\
$\vec{n}$ & Gaussian noise during optimization\\
$\vec{\mu}$ & Mean of the pruning vectors $\vec{v}$ during optimization\\
$\sigma$ & Standard deviation in Gaussian distribution\\
$\mathcal{N}(\vec{\mu},\sigma)$ & Gaussian distribution in sampling $\vec{v}$\\
$\rho$ & Regularization coefficient\\
$\alpha$ & Learning rate\\
$constraint$ & Target resource constraint\\
\hline
\end{tabular}}
\end{center}
\end{table}
\setlength{\tabcolsep}{1.4pt}

\section{Methodology}
\label{sec:method}
To solve the pruning problem in three dimensions, we define a unique mapping from the pruning vectors to it corresponding pruned network structure, where the pruning vector $\vec{v}$ is defined as the number of channels $c$ in each layer, and the spatial size $s$ and the network depth $d$ of the pruned network\footnote{To normalize the pruning vector $\vec{v}$ and make it isometric, we divide each entry by its maximum possible value, \textit{i.e.,} $\vec{v}$ = $\{c_1/c_{1_{max}}, c_2/c_{2_{max}},$ $ ..., s/s_{max}, d/d_{max}\}$}:
\begin{align}
    \vec{v} = \{c,s,d\},
\end{align}

Then, we formulate the pruning multiple dimensions as optimizing the numerical values of the pruning vector. The objective of this optimization problem is to minimize the loss under a given resource constraint:
\begin{align}
& {\rm minimize} \ \mathcal{L}(w;\vec{v}) \\
& {\rm subject \ to} \ \ \mathcal{C} (\vec{v}) < constraint
\end{align}
where $\mathcal{L}$ refers to the loss function of the neural network, it is a function of the weight parameters $w$, and is inexplicitly influenced by the pruning vector $\vec{v}$. The computational cost $\mathcal{C}$ is a function of the pruning vector $\vec{v}$ and $constraint$ denotes the target resource constraint (i.e., FLOPs or latency). We merge the resource constraint as a regularization term in the error function $\mathcal{E}$ as follows:
\begin{align}
    \mathcal{E}(w;\vec{v}) = \mathcal{L}(w;\vec{v}) + \rho ||\mathcal{C}(\vec{v})-constraint||^2,
\end{align}
where $\rho$ is the positive regularization coefficient. Consequently, the goal is to find the architecture pruning vector $v$ that minimizes the error function $\mathcal{E}$:
\begin{align}
   \vec{v} = \mathop{argmin}_{\vec{v}}(\mathcal{E})
\end{align}
However, it is impractical to use the exhaust algorithm to train every pruning network from scratch to obtain and compare their error $\mathcal{E}$. Instead, we proposed to directly update the numerical values in the pruning vector using gradient estimated from $\mathcal{E}$. Since there is no explicit function between the error $\mathcal{E}$ and the pruning vector $\vec{v}$, to tackle this, we propose to use the self-adapted gradient estimation method to approximate the steepest gradient descent direction of the pruning vector that minimizes the error. Further, we customize a weight sharing mechanism to provide weights for the error inference and enhance the optimization efficiency. We detail our gradient estimation method and the weight sharing mechanism as follows. Some notations are given in Table~\ref{table:notation}.

\subsection{Stochastic Gradient Estimation for Pruning Vector}
\label{sec:gradient_estimation}
Since the error function $\mathcal{E}$ is naturally non-differentiable with respect to the pruning vectors, we propose gradient estimation to approximate the underlying gradients. 
Inspired by~\cite{rechenberg1994evolutionsstrategie, yi2009stochastic, sehnke2010parameter}, we conduct the gradient estimation by relaxing the objective error function $\mathcal{E}$ to the expectation of $\mathcal{E}$ with pruning vector $\vec{v}$ obeying a distribution $p_{\theta}(\vec{v})$ .
\begin{align}
    \mathcal{E}(\vec{v}) \approx \mathbb{E}_{\vec{v} \sim p_{\theta}(\vec{v})}\mathcal{E}(\vec{v}),
\end{align}
where the distribution $p_{\theta}(\vec{v})$ is parameterized by $\theta$. To simplify the expression, we omit the $w$ in the term $\mathcal{E}(w;\vec{v})$.
Instead of learning the probability distribution $p_{\theta}(\vec{v})$ as a complex function of $\vec{v}$, we define $p_{\theta}(\vec{v})$ to be an isotropic multivariate gaussian distribution, to meet relaxation purpose in obtaining precise gradient estimation \textit{w.r.t.} the pruning vector $\vec{v}$:
\begin{align}
p_{\mu}(\vec{v})\sim \mathcal{N}(\vec{\mu},\sigma),
\end{align}
with the mean $\vec{\mu}$ being the current pruning vector and the deviation set to $\sigma$.
Consequently, the approximation to the objective function can be written as:
\begin{align}
	\mathcal{E}(\vec{\mu}) \approx \mathbb{E}_{\vec{v} \sim \mathcal{N}(\vec{\mu},\sigma)}\mathcal{E}(\vec{v}) = \mathbb{E}_{\vec{n} \sim \mathcal{N}(\vec{0},\sigma)}\mathcal{E}(\vec{\mu}+\vec{n}),
\end{align}
where $\vec{n}$ denotes the random multidimensional Gaussian noise added to the current pruning vector. This approximation holds because the neural network is \textit{Lipschitz} continuous with respect to $\vec{v}$~\cite{fazlyab2019efficient, scaman2018lipschitz, gouk2021regularisation}, i.e., there exists a positive real constant $K$ such that:
\begin{align}
\label{eq:lipschitz}
||\mathcal{E}(\vec{\mu}+\vec{n}) -  \mathcal{E}(\vec{\mu})|| < K||\vec{n}||.
\end{align}
In pruning vector optimization, $\vec{n}$ is confined to small variation values, i.e., $||\vec{n}|| < \epsilon$.
Therefore, the alteration in the error function is bounded within $K\epsilon$. In experiments, we observe the constant $K$ to be diminutive within a local region around the current pruning vector and thus the variation is tolerable.

\begin{figure*}[t]
\centering
\includegraphics[width=0.9\linewidth]{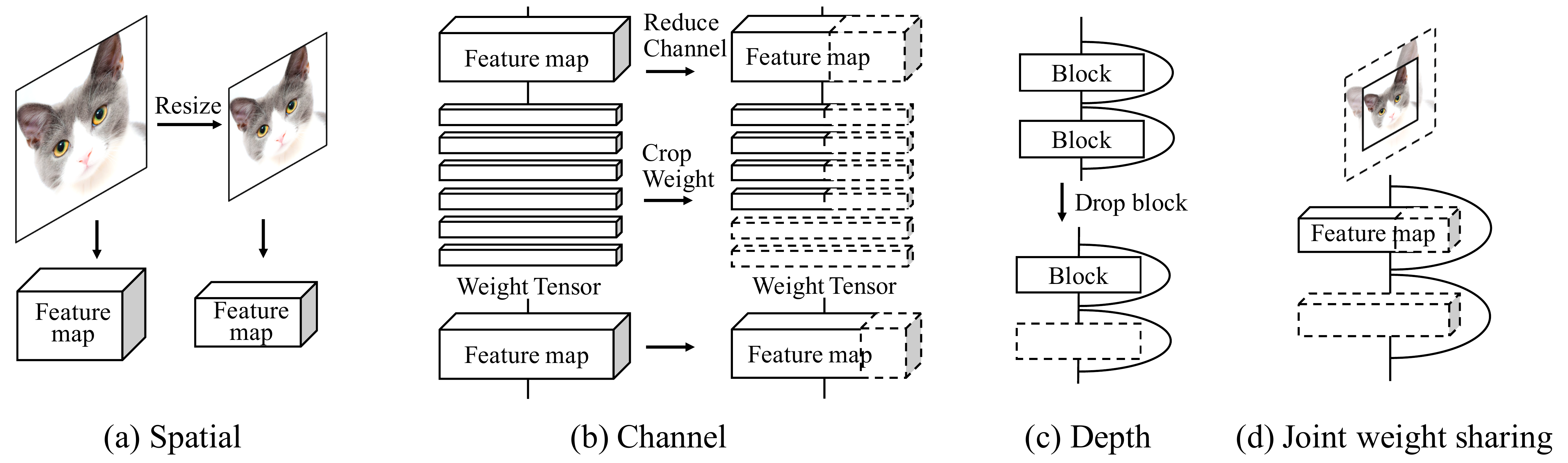}
\caption{The weight-sharing mechanism.}
\label{fig:weight_share}
\end{figure*}

With this approximation to the error function, we can derive the estimated gradient following the commonly adopted {\em log likelihood trick} in reinforcement learning:
\begin{align}
     &\nabla_{\vec{\mu}} \mathbb{E}_{\vec{v} \sim \mathcal{N}(\vec{\mu},\sigma)}\mathcal{E}(\vec{v})  = \mathbb{E}_{\vec{v} \sim \mathcal{N}(\vec{\mu},\sigma)} [\mathcal{E}(\vec{v}) \nabla_{\vec{\mu}} \log(p(\vec{v};\vec{\mu}))] \\
     & = \mathbb{E}_{\vec{v} \sim \mathcal{N}(\vec{\mu},\sigma)} [\mathcal{E}(\vec{v}) \nabla_{\vec{\mu}} (\log(\frac{1}{\sigma \sqrt{2\pi}}e^{-\frac{1}{2}(\frac{\vec{v}-\vec{\mu}}{\sigma})^2}))]\\
     & = \mathbb{E}_{\vec{v} \sim \mathcal{N}(\vec{\mu},\sigma)} [\mathcal{E}(\vec{v})\frac{\vec{v}-\vec{\mu}}{\sigma^2}] \\
     & = \frac{1}{\sigma^2} \mathbb{E}_{\vec{n} \sim \mathcal{N}(\vec{0},\sigma)} [\mathcal{E}(\vec{\mu}+\vec{n})\vec{n}].
\end{align}
Thus, the gradient of $\mathcal{E}(\vec{v})$ can be estimated by the pruning vector $\vec{v}$ stochastically sampled around the current vector $\vec{\mu}$ under a gaussian distribution $\vec{n}\sim \mathcal{N}(\vec{0},\sigma)$:
\begin{align}
\nabla_{\vec{\mu}} \mathbb{E}_{\vec{v} \sim \mathcal{N}(\vec{\mu},\sigma)}\mathcal{E}(\vec{v}) \approx \frac{1}{\sigma^2} \sum_{i=1}^M \mathcal{E}(\vec{\mu}+\vec{n}_i)\vec{n}_i,
\end{align}
where $M$ is the total number of samples. An intuitive interpretation of this gradient estimation approach is: by weighting the variation directions with the corresponding error, the gradient direction can be approximated with the variation direction that has the lower expected error.
Then the pruning vector is updated with the estimated gradient:
\begin{align}
\vec{\mu}' = \vec{\mu} - \alpha \nabla_{\vec{\mu}} \mathcal{E}(\vec{v}),
\end{align}
where $\alpha$ denotes the update rate, and the geometric meaning of $\vec{\mu}$ is the center pruning vector for adding Gaussian variations. The resulting algorithm iteratively executes the following two steps: (1) stochastic sampling of the architecture pruning vectors and evaluate the corresponding errors; (2) integrating the evaluations to estimate the gradient and update the pruning vector. 

\subsection{Weight Sharing}
\label{sec:weight_sharing}
In the aforementioned stochastic gradient estimation paradigm, one crucial issue that remains unsolved is how to obtain the error evaluation of the pruning vector.
Recall that the error function $ \mathcal{E}(w;\vec{v})$ is an explicit function of the weights $w$ associated with the corresponding pruning vector $\vec{v}$.
To evaluate the pruning vector, the weights in the corresponding pruned network need to be trained. However, it would be too computationally prohibitive to train each pruned network from scratch. To address this, we customize a parameter-sharing mechanism to share the weights for different architectures varying in the channel, spatial and depth configurations. 

The parameter-sharing mechanism is illustrated in Figure~\ref{fig:weight_share}. For spatial dimension, decreasing the input image resolution will reduce the feature map size in the network, as Figure~\ref{fig:weight_share}(a). This does not require any modification in the weight kernels and thus all the network weights are shared when adjusting the spatial resolution. To prune the channels, we follow~\cite{liu2019metapruning} to keep the first $c$ channels of feature maps in the original network and crop the weight tensors correspondingly, as shown in Figure~\ref{fig:weight_share}(b). To reduce depth, we adopt block dropping. That is keeping the first $d$ blocks and skipping the rest of the blocks, as Figure~\ref{fig:weight_share}(c). Combining these three dimensions, we come up with the joint parameter-sharing mechanism as illustrated in Figure~\ref{fig:weight_share}(d). With this mechanism, we only need to train one set of weights in the weight-shared network and evaluate the error of different pruning vectors with corresponding weights cropped from this network.

\subsection{Alternatively Update Weights and Pruning Vectors}
\label{sec:update}
For training the weights in the shared network, the optimization goal is set to minimize the error concerning the expectation of a bunch of pruning vectors. The objective is defined as:
\begin{align}
   \min_w \mathbb{E}_{\vec{v} \sim \mathcal{N}(\vec{\mu},\sigma)}\mathcal{E}(w; \vec{v}) 
\end{align}

To optimize this objective function, we inject noise to the pruning vectors during weight training. In each weight training iteration, the pruning vectors $\vec{v}$ is randomly sampled from $\mathcal{N}(\vec{\mu},\sigma)$. To compute the loss and gradients, we crop the weights in the weight-sharing matrix to match the size of the channels / depth specified in the pruning vector. Then we forward the particular batch of images, and calculate the loss and gradients. We use the gradients to update the weights that are used in the forward pass, other weights in the weight-sharing matrix remain intact.

After training the weights for one epoch, we use the stochastic gradient estimation to estimate the gradient \textit{w.r.t.} the $\vec{\mu}$ and take a step in the $\vec{\mu}$ variable space to decline the error:
\begin{align}
   \min_{\vec{\mu}} \mathbb{E}_{\vec{v} \sim \mathcal{N}(\vec{\mu},\sigma)}\mathcal{E}(w; \vec{v}) 
\end{align}
We alternatively optimize the weights and the architecture pruning vectors and meanwhile decrease the deviation $\sigma$ till convergence. With the error function $\mathcal{E}$ well approximated by its smoothed version $\mathbb{E}_{\vec{v} \sim \mathcal{N}(\vec{\mu},\sigma)}\mathcal{E}(w; \vec{v})$, we can obtain the optimized pruning vector to be the mean of the pruning vector distributions (i.e., $\vec{\mu}$) of the final model. The optimization pipeline is detailed in Algorithm~\ref{alg:1}.

\begin{algorithm}[h]
\caption{Joint Multi-Dimension Pruning}
\label{alg:1}
\textbf{Input:} Learning rate $\alpha$, standard deviation of gaussian distribution $\sigma$, initial mean $\vec{\mu}_0$, pruning vector $\vec{v}$, initial weight: $w$, error function: $\mathcal{E}$, number of total iterations: $K$, number of weight update iterations: $N$, number of configuration samples: $M$ .\\
\textbf{Output}: Optimized configuration $\mu^*$\\
\begin{algorithmic}[1]
\FOR{$t = 0:K$} 
\FOR{$i = 0:N$}
\STATE  $ \min_w \mathbb{E}_{\vec{v} \sim \mathcal{N}(\vec{\mu},\sigma)}\mathcal{E}(w; \vec{v})$  
\ENDFOR
\FOR{$j = 0:M$} 
\STATE $\mathcal{E}_j = \mathcal{E}(\vec{\mu}_t+\vec{n}_j)$, $\vec{n}_j\sim \mathcal{N}(\vec{0},\sigma)$
\ENDFOR
\STATE $\mu_{t+1} = \mu_t - \alpha \frac{1}{\sigma^2} \sum_{i=1}^M \mathcal{E}(\vec{\mu}_t+\vec{n}_j) n_j$
\ENDFOR
\STATE return $\mu^*$
\end{algorithmic}
\end{algorithm}

\section{Experiments}
In this section, we begin by introducing our experiment settings. Then, we explain the pruning vector settings. Thirdly, we show the comparison between our method and the state-of-the-arts on various architectures and under different resource constraint metrics. Lastly, we visualize the pruned networks obtained by our algorithm and the pruning vector convergence curve during optimization as well as provide ablation studies.

\subsection{Experimental Settings}
\label{sec:exp_setting}
\noindent\textbf{Dataset} All experiments are conducted on the ImageNet 2012 classification dataset~\cite{russakovsky2015imagenet}. For optimizing the pruning vector, we randomly split the original training set into two subsets: 50,000 images for validation (50 images for each class) and use the rest as the training set. After the architecture hyperparameters are optimized, we train the corresponding architecture from scratch using the original training/validation dataset split, following the practice in~\cite{cai2018proxylessnas,wu2018fbnet}.

\noindent\textbf{Training}
We choose the overall number of iterations as one-fourth of the total iterations in training the MobileNet~\cite{howard2017mobilenets,sandler2018mobilenetv2} from scratch, following the practice in MetaPruning~\cite{liu2019metapruning}. We found that prolonging the training for longer does not bring further benefits in the searched network accuracy. In detail, we alternatively optimize the pruning vector and weights with an outer loop of 100 iterations.
In each inner loop, we train the weights for 2000 iterations for MobileNet V1/V2 and 1000 iterations for ResNet with a batch size of 256. So that the weights can learn to adapt for multiple architecture vectors sampled in the local region of the current pruning vector. We use the SGD optimizer with momentum 0.9, weight decay 1e-5 and learning rate 0.01 for weight training. Then we update the pruning vector for 20 iterations with the estimated gradients. In gradient estimation,  the number of samples $M$ is set to 100 for each iteration, the deviation $\sigma$ and the update rate $\alpha$ are initialized with 1.25 and 5 respectively, and linearly decay to 0.25 and 0, respectively, with the outer loop. 

\noindent\textbf{Constraints}
The experiments are carried out under the FLOPs, GPU latency as well as CPU latency constraints. For the latency constraints, we follow the practice in FBNet~\cite{wu2018fbnet} to build a latency lookup table for a layer with different configurations and obtain the total latency of the network by summing up the latency of all layers. In our experiments, we estimate GPU latency on a single GTX 1080Ti with a batch size of 256, and it cost $\sim$ 0.5 GPU days to build the lookup table. For CPU latency, we use the lookup table released by ChamNet~\cite{dai2018chamnet} for a fair comparison. As ChamNet has a very sparse lookup table, we use the Gaussian process to predict the missing latency values as \cite{dai2018chamnet} did for energy estimation. The squared error of the prediction is within 0.5\%. 

\subsection{Details in Pruning Vector Settings}
\label{sec:configuration_setting}
For MobileNet V1, we set the pruning vector ($v$) as $v=\{c_1,c_2,…c_n, s, d\}$, where $c_i$ are the numbers of output channels in layer $i$, and $n$ denotes total number of layers. $s$ is the spatial size of the input image and $d$ is the network depth. Then we normalize each entry in $v$ to [0,1] by dividing with the maximum value of that entry for gradient computation and vector update. When constructing the pruned network based on the pruning vector, we time the normalized value by the maximum value of each entry and round the channel number, depth and spatial resolutions to integers. MobileNet V1 is a network without shortcuts, we only drop the block that has the same input and output channel number. We store the weight matrix as 1.5$\times$ channels in the baseline network and initialize the pruning vector as the configuration of the original baseline network. After obtaining the optimal configuration, the pruned network is constructed through pruning each layer’s input channel equaling to the previous layer’s output channel, specified by $c_i$. 

Similar to MobileNet V1, the pruning vector ($v$) for MobileNet V2, MobileNet V3 and ResNet is defined as $v=\{c_1,c_2,…c_n, s, d\}$. Since MobileNet V2/V3 and  ResNet contain the shortcut structure, we confine the layers connected with identity shortcuts to have the same number of output channels. For pruning the depth dimension, we drop the last blocks and keep the first $d$ blocks.

\begin{figure}[t]
\centering
\includegraphics[width=0.95\linewidth]{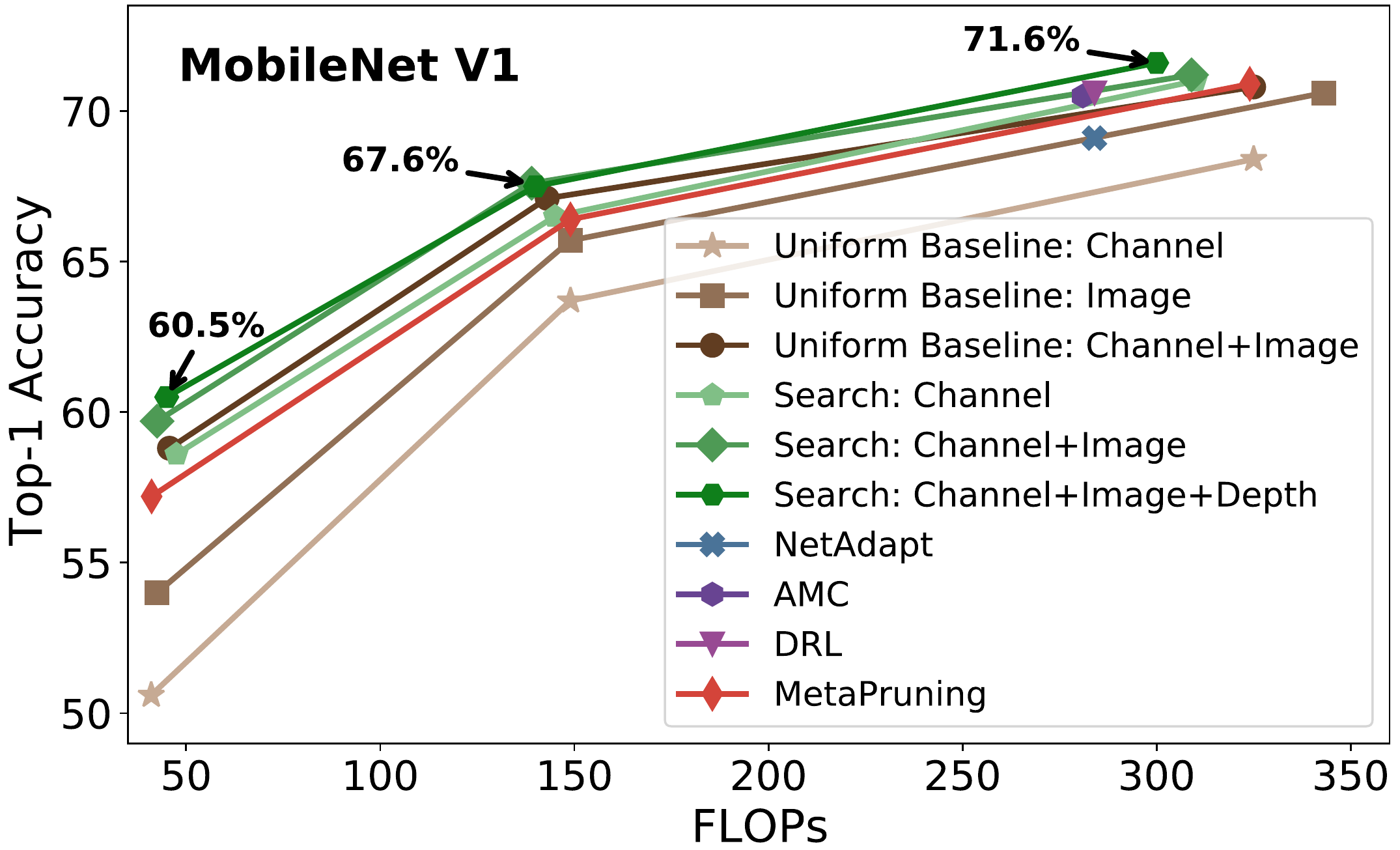}
\caption{Comparison of the proposed method with uniform baselines and state-of-the-art AutoML-based pruning methods (Netadapt~\cite{yang2018netadapt}, AMC~\cite{he2018amc}, Pruning via DRL~\cite{chen2019storage} and MetaPruning~\cite{liu2019metapruning}) on MobileNet V1~\cite{howard2017mobilenets}. The baseline is obtained by uniformly rescale channel/image/channel+image by a fixed ratio.}
\label{fig:mbv1_FLOPs}
\end{figure}

\setlength{\tabcolsep}{1pt}
\begin{table}
\begin{center}
\caption{Comparison between the proposed method with MobileNet V1~\cite{howard2017mobilenets} uniform baselines, under the latency constraints. Reported GPU latency is the run-time of the corresponding network on a single 1080Ti with a batch size of 256.}
\label{table:mbv1_latency}
\resizebox{.46\textwidth}{!}{
\begin{tabular}{ccccccccc}
\noalign{\smallskip}
\hline
\hline
\multicolumn{3}{c}{\textbf{Uniform Baseline \ \ \ }} & \multicolumn{2}{c}{\textbf{Proposed Method}} \\
\hline
\multicolumn{3}{c}{\textbf{Rescale: Channel \ \ \ }} & \multicolumn{2}{c}{\textbf{Optimize: Channel+Image}} \\
\hline
Ratio \ \ \ &  \ \ Top1-Acc \ \ & \ \ Latency \ \ & \ \ Top1-Acc \ \ & \ \ Latency \ \ \\
\hline
1$\times$ & 70.6\% & 7.806ms & -- & --  \\
\hline
0.75$\times$ & 68.4\% & 5.620ms & \textbf{72.3\%} & 5.617ms\\
\hline
0.5$\times$ & 63.7\% & 3.998ms & \textbf{70.6\%} & 3.972ms\\
\hline
0.25$\times$ & 50.6\% & 2.266ms & \textbf{66.8\%} & 2.231ms\\
\hline
\hline
\end{tabular}}
\end{center}
\end{table}
\setlength{\tabcolsep}{1.4pt}

\begin{figure}[t]
\centering
\includegraphics[width=0.95\linewidth]{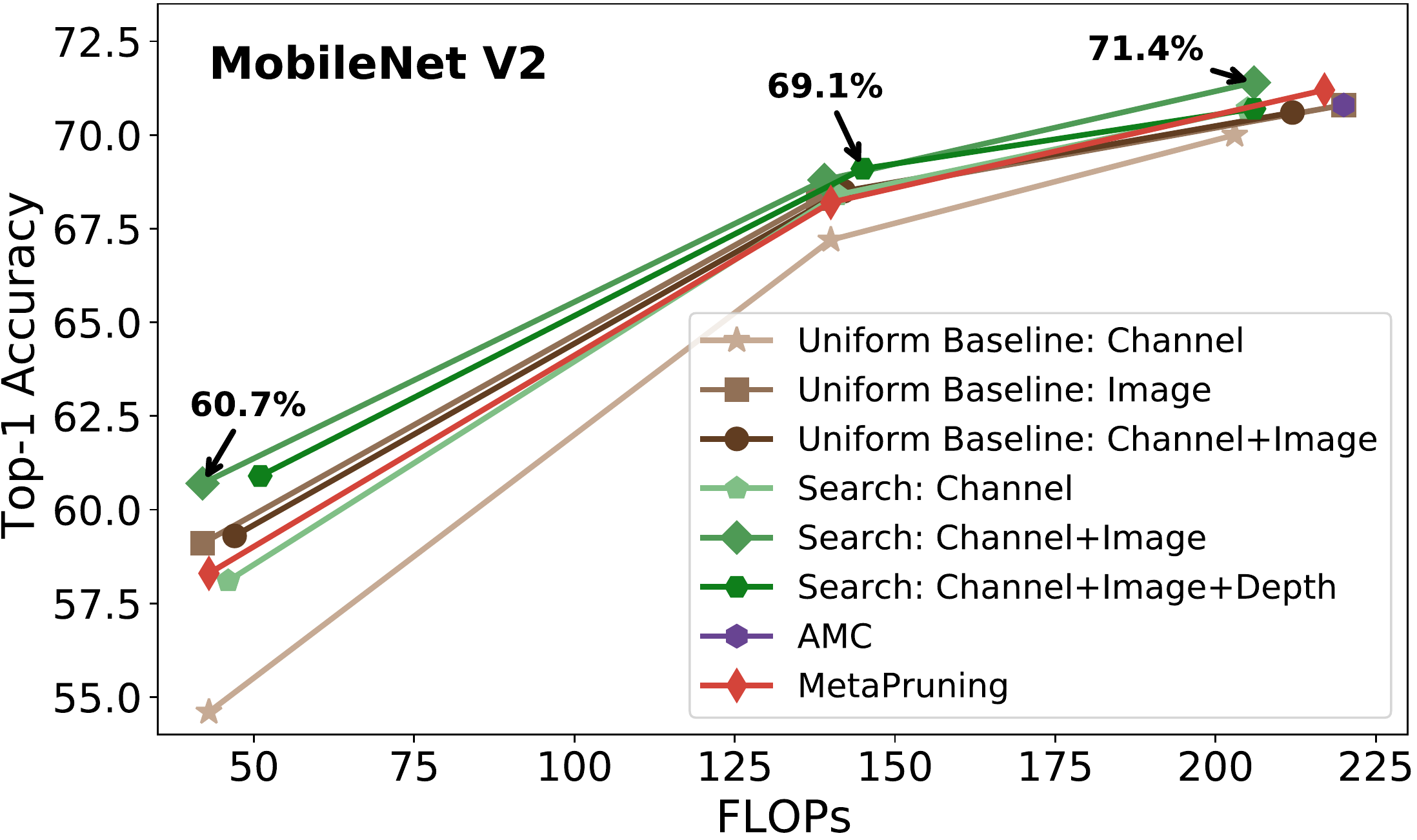}
\caption{Comparison between the proposed method with uniform baselines and state-of-the-art AutoML based pruning methods (AMC~\cite{he2018amc} and MetaPruning~\cite{liu2019metapruning}) on MobileNet V2~\cite{sandler2018mobilenetv2}.}
\label{fig:mbv2_FLOPs}
\end{figure}

\setlength{\tabcolsep}{1pt}
\begin{table}[t]
\begin{center}
\caption{Comparison between the proposed method with the MobileNet V2~\cite{sandler2018mobilenetv2} uniform baselines, under the GPU latency constraint, which is measured on one 1080Ti GPU with batch size 256.}
\label{table:mbv2_gpu_latency}
\resizebox{.48\textwidth}{!}{
\begin{tabular}{ccccccccc}
\noalign{\smallskip}
\hline
\hline
\multicolumn{2}{c}{\textbf{\ \ \ \ Uniform Baseline }} & \multicolumn{4}{c}{\textbf{Proposed Method}} \\
\hline
\multicolumn{2}{c}{\textbf{Rescale: Channel}} & \multicolumn{2}{c}{\textbf{Optimize: C+I}} & \multicolumn{2}{c}{\textbf{Optimize: C+I+D}} \\
\hline
  \ \ \ \ Acc & Latency & Acc & Latency & Acc & Latency\\
\hline
  \ \ \ \ 70.0\% \ & \ \ 7.36ms \ & \ 72.3\% \ \ & \ \ 7.34ms \ \ & \ \textbf{72.4\%} \ \ & \ \ 7.17ms \ \ \\
\hline
  \ \ \ \ 67.2\% \ & \ \ 5.97ms \ & \ 71.2\% \ \ & \ \ 5.93ms \ \ & \ \textbf{71.7\%} \ \ & \ \ 5.90ms \ \ \\
\hline
  \ \ \ \ 54.6\% \ & \ \ 4.22ms \ & \ 67.7\% \ \ & \ \ 4.10ms \ \ & \ \textbf{68.7\%} \ \ & \ \ 3.98ms \ \ \\
\hline
\hline
\end{tabular}}
\end{center}
\end{table}
\setlength{\tabcolsep}{1.4pt}

\begin{table}[h]
\begin{center}
\caption{Comparison of the proposed method jointly optimizing three dimensions (\textit{i.e.}, channel, spatial and depth) on MobileNet V3~\cite{howard2019searching} (Large) structure with uniform baselines and state-of-the-art gradient-based neural architecture search methods.}
\label{table:mbv3_FLOPs}
\resizebox{.46\textwidth}{!}{
\begin{tabular}{cccccccc}
\noalign{\smallskip}
\hline
\hline
\multirowcell{3}{Uniform  \ \ \ \\ Baseline  \ \ \ } 
&  & Rescale Dimension & Top1-Acc & FLOPs \ \\
\cline{2-5}
& & No Rescale &75.2\% & 219M \\
\cline{2-5}
& & {Channel + Input image} & 75.6\% & 350M \\
\hline
\hline
\multirowcell{6}{State-of- \ \ \ \\ the-arts \ \ \ } 
& \multicolumn{2}{c}{Method} & Top1-Acc & FLOPs \\
\cline{2-5}
&\multicolumn{2}{c}{ProxylessNAS~\cite{cai2018proxylessnas}} & 74.8\% & 465M\\
&\multicolumn{2}{c}{EfficientNet~\cite{tan2019efficientnet}}  & 77.1\% & 390M\\
&\multicolumn{2}{c}{FB-Net-C~\cite{wu2018fbnet}}  & 75.0\% & 375M\\
&\multicolumn{2}{c}{Single-Path NAS~\cite{stamoulis2019single}}  & 74.9\% & 334M\\
&\multicolumn{2}{c}{Single-Path One Shot~\cite{guo2019single}}  & 74.7\% & 328M\\
\hline
\hline
\multirowcell{2}{Proposed \ \ \ \\ Method \ \ \ } 
& \multicolumn{2}{c}{Optimization Space} & Top1-Acc & FLOPs \\
\cline{2-5}
& \multicolumn{2}{c}{Channel + Input image + Depth} & 77.3\% & 344M\\
\hline
\hline
\end{tabular}}
\end{center}
\end{table}
\setlength{\tabcolsep}{1.4pt}

\begin{table}[h]
\begin{center}
\caption{Comparison between the proposed method on MobileNet V3~\cite{howard2019searching} with the state-of-the-art NAS methods under the GPU latency constraint. \textit{Gds} is abbreviation for GPU days.}
\label{table:mbv3_gpu_latency}
\setlength{\tabcolsep}{3pt}
\resizebox{.44\textwidth}{!}{
\begin{tabular}{ccccccccc}
\noalign{\smallskip}
\hline
\hline
& FB-Net-C & ProxylessNAS & SPOS & \multirow{2}{*}{Ours} \\
& \cite{wu2018fbnet} & \cite{cai2018proxylessnas} & \cite{guo2019single} & \\
\hline
Accuracy & 74.9\%  & 74.8\% & 75.3\% & \textbf{75.6\%} \\
Latency & 19ms & 22ms & 22ms & 20ms\\
Total Time Cost & 36 \textit{Gds} & 31 \textit{Gds} & 29 \textit{Gds} & 22 \textit{Gds}\\
\hline
\hline
\end{tabular}}
\end{center}
\end{table}

\subsection{Comparisons with State-of-the-arts}
\label{sec:sota}

To verify the effectiveness of the proposed JointPruning, we compare our method with the uniform baselines, traditional pruning methods as well as the AutoML-based model adaptation methods. Our experiments are conducted on MobileNet V1/V2/V3~\cite{howard2017mobilenets,sandler2018mobilenetv2,howard2019searching} and ResNet~\cite{he2016deep} backbones.

\textbf{MobileNet V1\&V2.}
\label{subsec:mbv1}
MobileNets~\cite{howard2017mobilenets,sandler2018mobilenetv2,howard2019searching} are already compact networks, many recent pruning algorithms focus on these structures for finding compact pruned networks with practical significance.

The uniform baseline groups in Fig.~\ref{fig:mbv1_FLOPs} and Fig.~\ref{fig:mbv2_FLOPs} show that rescaling channel and spatial dimension by the same ratio can achieve higher accuracy than merely rescale channel or spatial under the same FLOPs constraint, suggesting the importance of jointly dealing with multi-dimensions. 
Then, for jointly optimizing the layer-wise channel numbers and the spatial size, the proposed JointPruning can automatically capture the delicate linkage between spatial pruning and channel-wise pruning ratio adjustments, which outputs optimized pruned networks with higher accuracy than state-of-the-art pruning methods focusing on channels only. We found that the pruned network architecture discovered by JointPruning captures meaningful patterns by preserving more channels for the shallower layers when the input resolution is small. This pruning scheme helps the network encodes more channel information to compensate for the information loss in the input size reduction.
When further considering the depth dimension, JointPruning obtains even better accuracy, especially under the extremely small FLOPs constraint.

Besides the FLOPs constraint, we extensively study the behavior of JointPruning under direct metrics like GPU latency and CPU latency, which has practical importance when deploying a model with a requirement on real-time computation speed.
Given various latency constraints, JointPruning can automatically utilize the hardware characteristics and automatically discover suitable pruned networks correspondingly. For GPUs with high parallel computing ability, the searched pruned networks are more likely to have more channels while less depth to fully utilize the degree of parallelism. The ability of JointPruning in flexibly learning the trade-off between different dimensions consistently improves the accuracy of the pruned network and surpasses the uniform pruning by a remarkable margin when targeting at the latency constraints. 
As shown in Table~\ref{table:mbv1_latency}, on MobileNet V1 JointPruning with the channel and spatial dimension (C+S) achieves more than 3.9\% accuracy enhancements. Table~\ref{table:mbv2_gpu_latency} shows that, on MobileNet V2, JointPruning comprehensively considers channel, spatial and depth (C+S+D) dimensions, which further boosts the accuracy by 2.4$\sim$4.3\%.

Compared to the state-of-the-art model adaptation method ChamNet~\cite{dai2018chamnet} building atop the Gaussian process, which is computation-costly because it needs to train 240 architectures from scratch to build the performance prediction curve. Instead, JointPurning with stochastic gradient estimation and weight sharing mechanism enjoys high optimization efficiency, which only takes 40 GPU hours for finding an optimal pruned network. As shown in Table~\ref{table:mbv2_cpu_latency}, JointPruning achieves comparable or higher accuracy compared to ChamNet, but with much lower optimization time cost. Compared to MetaPruning~\cite{liu2019metapruning} which utilized a two-step framework of training a PruningNet and searching for the Pruned Network, JointPruning only requires one-step to joint optimize the pruning vectors in three dimensions and the weight parameters, which is more efficient. Moreover, JointPruning assumes a Gaussian distribution in the pruned network sampling, involving no tuning of RL hyperparameters compared to AMC~\cite{he2018amc}. JointPruning also achieves competitive results compared to MetaPruning and AMC on the channel dimension solely and has the additional ability of dealing with multiple dimensions beyond the channel.

\setlength{\tabcolsep}{1pt}
\begin{table}
\begin{center}
\caption{This table compares the Top-1 accuracy of the proposed method with the ChamNet~\cite{dai2018chamnet}, using the same CPU latency lookup table provided by \cite{dai2018chamnet}.}
\label{table:mbv2_cpu_latency}
\resizebox{.44\textwidth}{!}{
\begin{tabular}{ccccccccc}
\noalign{\smallskip}
\hline
\hline
 & \multicolumn{2}{c}{\textbf{\ \ \ \ ChamNet }} & \multicolumn{2}{c}{\textbf{Proposed Method}} \\
\hline
\multirow{4}{*}{Performance}
 & \ \ \ \ Acc & Latency & Acc & Latency\\
\cline{2-5}
 & \ \ \ \ \textbf{71.9\%} \ & \ \ 15.0ms \ & \ 71.8\% \ \ & \ \ 14.8ms \ \ \\
\cline{2-5}
 & \ \ \ \ 69.0\% \ & \ \ 10.0ms \ & \ \textbf{69.0\%} \ \ & \ \ 9.9ms \ \ \\
\cline{2-5}
 & \ \ \ \ 64.1\% \ & \ \ 6.1ms \ & \ \textbf{66.4\%} \ \ & \ \ 6.0ms \ \ \\
\hline
Total Time Cost & \multicolumn{2}{c}{5760 GPU days} & \multicolumn{2}{c}{\textbf{3 $\times$ 40 GPU days}} \\
\hline
\hline
\end{tabular}}
\end{center}
\end{table}
\setlength{\tabcolsep}{1.4pt}

\setlength{\tabcolsep}{1pt}
\begin{table}[!tp]
\begin{center}
\caption{This table compares the Top-1 accuracy of the proposed method with uniform baselines and channel pruning methods on ResNet-50~\cite{he2016deep}. The accuracy of PFEC~\cite{pruningfilters} is quoted from \cite{huang2018data}. Other results are quoted from the original paper.}
\label{table:resnet_FLOPs}
\resizebox{.425\textwidth}{!}{
\begin{tabular}{cccccccc}
\noalign{\smallskip}
\hline
\hline
\multicolumn{3}{c}{Method} & Top1-Acc & FLOPs \\
\hline
\multicolumn{3}{c}{PFEC~\cite{pruningfilters}} & 72.9\% & 3.1G\\
\multicolumn{3}{c}{ThiNet-70~\cite{luo2017thinet}} & 72.0\% & 2.9G\\
\multicolumn{3}{c}{C-SGD-70~\cite{ding2019centripetal}} & 75.3\% & 2.9G\\
\multicolumn{3}{c}{IENNP-91\%~\cite{molchanov2019importance}} &75.5\% & 2.7G\\
\hline
\multicolumn{3}{c}{SFP~\cite{he2018soft}} & 74.6\% & 2.4G \\
\multicolumn{3}{c}{FPGM~\cite{he2019filter}} &75.6\% & 2.4G \\
\multicolumn{3}{c}{VCNNP~\cite{zhao2019variational}} & 75.2\% & 2.4G \\
\multicolumn{3}{c}{AP(r= 0.5)~\cite{luo2018autopruner}} & 74.8\% & 2.3G \\
\multicolumn{3}{c}{IENNP-72\%~\cite{molchanov2019importance}} &74.5\% & 2.3G\\
\multicolumn{3}{c}{Slim~\cite{networkslimming}} & 74.9\% & 2.3G\\
\multicolumn{3}{c}{GAL-0.5~\cite{lin2019towards}} & 72.0\% & 2.3G \\
\multicolumn{3}{c}{GDP-0.7~\cite{lin2018accelerating}} & 72.6\% & 2.2G \\
\multicolumn{3}{c}{C-SGD-50~\cite{ding2019centripetal}} & 74.5\% & 2.1G\\
\multicolumn{3}{c}{ThiNet-50~\cite{luo2017thinet}} & 71.0\% & 2.1G\\
\multicolumn{3}{c}{DCP~\cite{zhuang2018discrimination}} & 74.9\% & 2.1G \\
\multicolumn{3}{c}{CP~\cite{he2017channelpruning}} & 73.3\% & 2.0G\\
\multicolumn{3}{c}{RRBP~\cite{zhou2019accelerate}} & 73.0\% &1.9G\\
\multicolumn{3}{c}{GDP-0.5~\cite{lin2018accelerating}} & 70.9\% & 1.6G \\
\hline
\multicolumn{3}{c}{IENNP-56\%~\cite{molchanov2019importance}} &71.7\% & 1.3G\\
\multicolumn{3}{c}{ThiNet-30~\cite{luo2017thinet}} & 68.4\% & 1.2G\\
\multicolumn{3}{c}{AP(r= 0.3)~\cite{luo2018autopruner}} & 73.1\% & 1.1G \\
\hline
\hline
\multirowcell{5}{ Uniform \\ Baseline } 
& Dimension \ & Rescale Ratio \  & Top1-Acc \ & FLOPs \ \\
\cline{2-5}
&& 1$\times$ &76.6\% & 4.1G \\
\cline{3-5}
& \multirow{3}{*}{Channel} 
& 0.85$\times$ & 76.0\% & 3.2G \\
&& 0.75$\times$ & 74.8\% & 2.3G \\
&& 0.5$\times$ & 72.0\% & 1.1G \\
\hline
\hline
\multirowcell{7}{Proposed \\ Method } 
& \multicolumn{2}{c}{Optimization Space} & Top1-Acc & FLOPs \\
\cline{2-5}
& \multicolumn{2}{c}{\multirowcell{3}{Layer-wise \\ Channel Numbers}}
& 76.2\% & 3.0G \\
&&& 75.6\% & 2.0G \\
&&& 73.4\% & 1.0G\\
\cline{2-5}
& \multicolumn{2}{c}{\multirowcell{3}{Layer-wise Channel \\ Numbers + Input image \\ + Depth}}
& \textbf{76.5\%} & 2.8G \\
&&& \textbf{76.0\%} & 1.9G \\
&&& \textbf{73.5\%} & 0.9G\\
\hline
\hline
\end{tabular}}
\end{center}
\end{table}
\setlength{\tabcolsep}{1.4pt}

\begin{table}
\caption{JointPruning vs. Random select on MobileNet V2 structure.}
\label{table:random_search}
\centering
\resizebox{.31\textwidth}{!}{
\begin{tabular}{ccccccccccc}
\hline
\hline
\noalign{\vspace{1.5pt}}
& Random select & Ours \\
\noalign{\vspace{1.5pt}}
\hline
\noalign{\vspace{1.5pt}}
Acc(\%) & 65.2 $\pm$ 2.5 & \textbf{69.1}\\
FLOPs \ \ & \ 146.5M $\pm$ 1.5M \ & \ \textbf{145M}\ \\
\noalign{\vspace{1.5pt}}
\hline
\noalign{\vspace{1.5pt}}
Acc(\%) & 68.2 $\pm$ 2.9  &  \textbf{71.7}\\
Latency \ \ & 5.985ms $\pm$ 0.015ms & \ \ \textbf{5.90ms}\ \ \\
\noalign{\vspace{1.5pt}}
\hline
\hline
\end{tabular}}
\end{table}

\textbf{MobileNet V3.} MobileNet V3~\cite{howard2019searching} is a state-of-the-art compact network. Based on this backbone structure, we further investigate the performance of the proposed JointPruning in finding the best ratio for scaling up the network. We define the maximum channel number as 1.5 $\times$ the baseline network. In Table~\ref{table:mbv3_FLOPs}, our proposed method discovers an architecture that achieves 77.3\% accuracy, surpassing the previous NAS methods. The training technique follows~\cite{howard2019searching}, with knowledge distillation~\cite{hinton2015distilling,shen2020meal}. When searching under the latency constraints, as shown in Table~\ref{table:mbv3_gpu_latency}, our method also obtains higher accuracy than the state-of-the-art NAS methods with a lower search time cost. This demonstrates the high efficiency of the proposed JointPruning method. Compared with other NAS methods like FB-Net~\cite{wu2018fbnet}, which model different channel choices as separate options, the proposed JointPruning method utilizes the continuity in channel numbers configurations and directly updates the numerical values with estimated gradients, which allows the algorithm to explore the search space more effectively.

\textbf{ResNet.}
\label{subsec:resnet}
We show JointPruning is also effective when targeting at channel dimension only. In comparison to the traditional channel pruning methods, as shown in Table~\ref{table:resnet_FLOPs}, JointPruning achieves better performance with less human participation. The accuracy enhancement mainly comes from JointPruning’s ability in pruning shortcuts. For traditional pruning with a layer-by-layer scheme, shortcuts will affect more than one layer which is tough to deal with. While JointPruning can effortlessly prune the shortcut by modeling the shortcut pruning as updating the numerical channel numbers. We further show that by incorporating the depth and spatial dimensions into the search space, we can harvest higher accuracy than pruning the channels solely given the same computational constraints. 

\textbf{Random method}
Compared to the random method which determines the configuration in three dimensions randomly, Table~\ref{table:random_search} shows that JointPurning is able to end up at an optimized minimum with high accuracy, and significantly outperforms the random scheme.

\subsection{Optimization Results Visualization}
\label{sec:visualization}
\begin{figure}[t]
\centering
\includegraphics[width=0.8\linewidth]{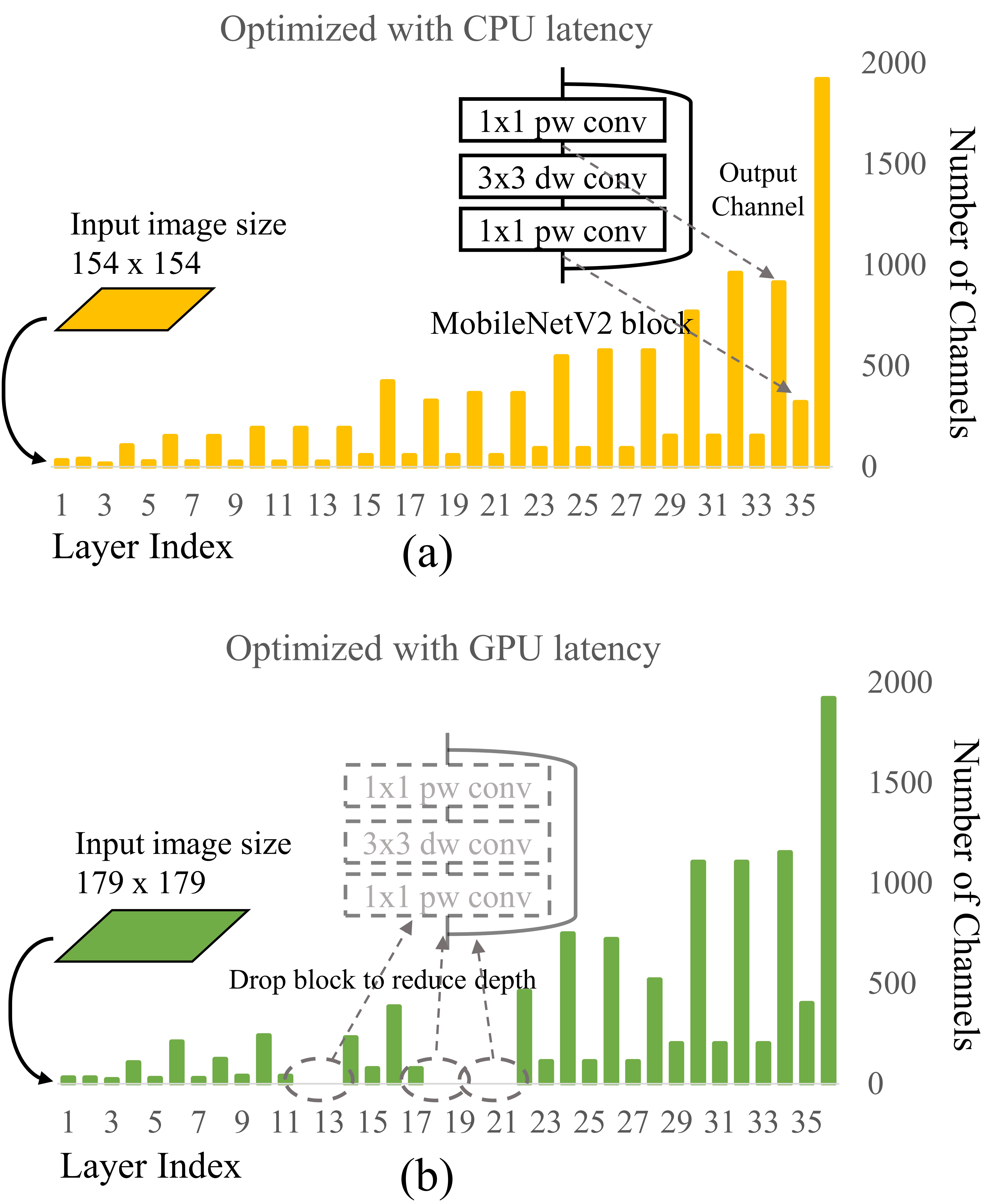}
\caption{This figure shows the architecture hyper-parameters (input image size, layer-wise channel numbers and depth) of  MobileNet V2 network structure optimized under (a) CPU latency constraint with latency = 10ms and (b) GPU latency constraint with latency = 5.62ms.}
\label{fig:visualization}
\end{figure}

In this section, we visualize two sets of pruning configurations for MobileNetV2 focusing on optimizing CPU and GPU latency, respectively. By incorporating the latency constraint into the error function $\mathcal{E}$, JointPruning is able to take advantage of hardware characteristics without knowing the underlying implementation details. As shown in Fig.~\ref{fig:visualization}(a), the final network optimized under the CPU latency constraint is deep with smaller spatial resolution. While for the GPU, shown in Fig.~\ref{fig:visualization}(b), the corresponding architecture discovered by JointPruning adopts large spatial resolution, more channels with fewer layers to fully utilize the parallel computing capability of GPUs. 

\begin{figure}[t]
\centering
\includegraphics[width=0.8\linewidth]{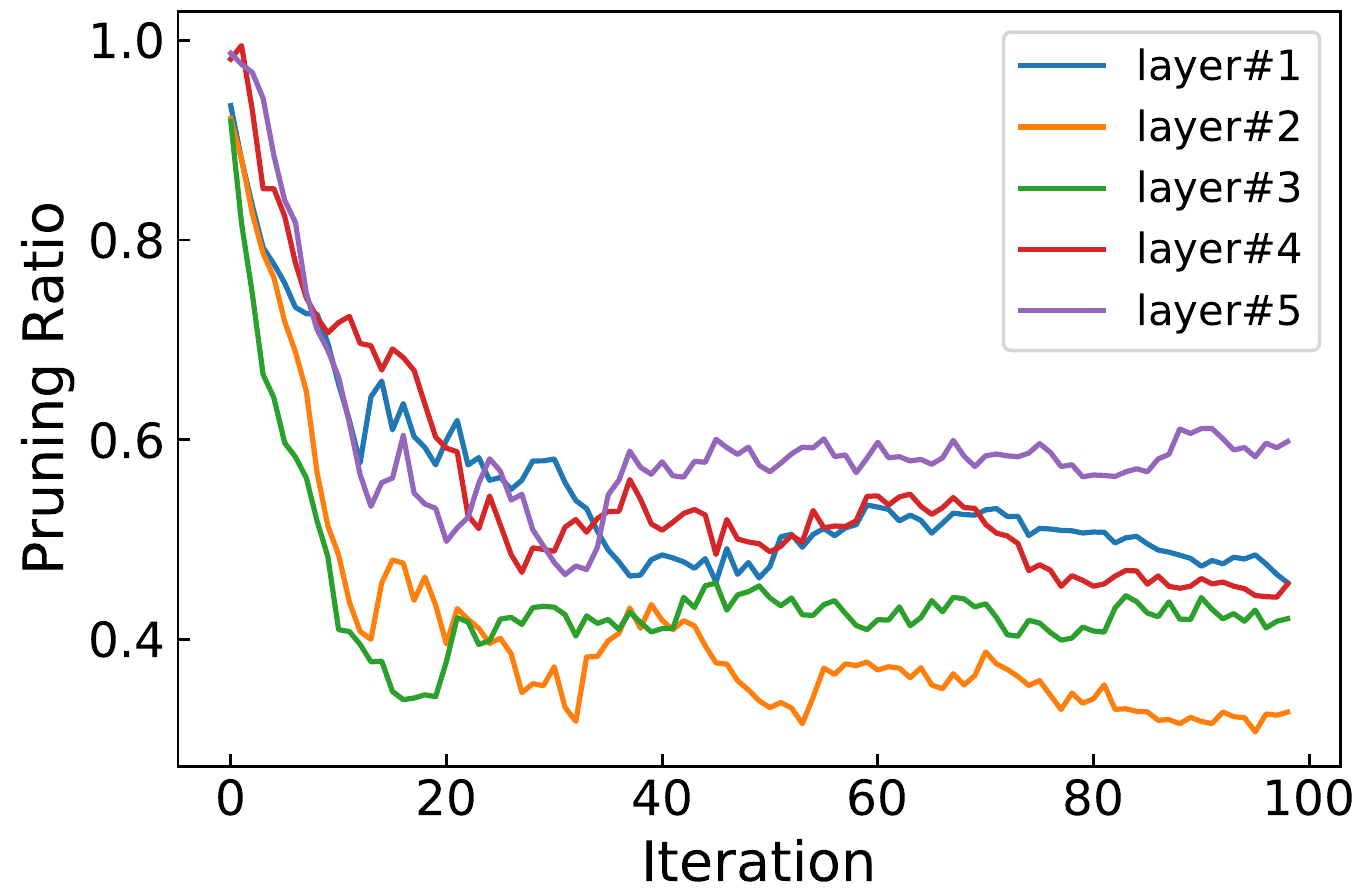}
\caption{In this figure, we visualize the pruning ratio convergence curve of the first five layers in pruning MobileNet v1~\cite{howard2017mobilenets}.}
\label{fig:pruning_curve}
\end{figure}

\subsection{Pruning Vector Convergence Curve Visualization}
\label{sec:pruning_curve}
We visualize the pruning ratio vs. iteration on MobileNet v1 (45M) in Fig.~\ref{fig:pruning_curve}. Each curve represents the trajectory of pruning ratio in each layer. It can be observed that after $\sim$75 iterations, the oscillation amplitude reduces to lower than 0.05 and the variance of pruning ratio tends to be stable. After training, the pruning vector in the multi-dimensional pruning converges to optimal pruning configuration.

\begin{table}[h]
\caption{The lower bound of $K$ in Eq.~\ref{eq:lipschitz}.}
\label{table:lipschitz}
\centering
\resizebox{.28\textwidth}{!}{
\begin{tabular}{ccccccccccc}
\hline  
\hline
\noalign{\vspace{1.5pt}}
MobileNet V1 & MobileNet V2 & ResNet \\
\noalign{\vspace{1.5pt}}
\hline
\noalign{\vspace{1.5pt}}
8.4628 & 7.3885 & 5.5479 \\
\noalign{\vspace{1.5pt}}
\hline
\hline
\end{tabular}}
\end{table}

\begin{table}[h]
\caption{Ablation study on the effect of the regularization hyperparameter $\rho$. Experiments are conducted with the ResNet-50 structure on ImageNet dataset.}
\label{table:rho}
\centering
\resizebox{0.42\textwidth}{!}{
\begin{tabular}{ccccccccccc}
\hline
\hline
\noalign{\vspace{1.5pt}}
{$\rho$} & {1\textit{e}-8} & {5\textit{e}-9} & {2\textit{e}-9} & {1\textit{e}-9} \\
\noalign{\vspace{1.5pt}}
\hline
\noalign{\vspace{1.5pt}}
{Target FLOPs} &  {1G}  & {1G} & {1G}  & {1G} \\
{Initial FLOPs} &  {4.1G}  & {4.1G} & {4.1G}  & {4.1G} \\
{Initial FLOPs Loss} &  {961}  & {240.25} & {38.44}  & {9.61} \\
{Final FLOPs} & {0.994G}  & {1.005G} & {1.020G}  & {1.035G} \\
{Final FLOPs Loss} &  {3.624\textit{e}-3}  & {7.33\textit{e}-4} & {1.541\textit{e}-3}  & {1.226\textit{e}-3} \\
{Top-1 Acc } & {72.59}  & {72.61} & {73.50}  & {73.11} \\
\noalign{\vspace{1.5pt}}
\hline
\hline
\end{tabular}}
\end{table}

\begin{table}[!h]
\caption{{Ablation study on the standard deviation $\sigma$ in updating the pruning vector. Experiments are conducted with the ResNet-50 structure on ImageNet dataset.}}
\label{table:sigma}
\centering
\resizebox{0.48\textwidth}{!}{
\begin{tabular}{ccccccccccc}
\hline
\hline
\noalign{\vspace{1.5pt}}
{Initial value of $\sigma$} & {0.05} & {0.02} & {0.015} & {0.0125} & {0.01} & {0.005} & {0.0025}\\
\noalign{\vspace{1.5pt}}
\hline
\noalign{\vspace{1.5pt}}
{Target FLOPs} & {950M}  & {950M} & {950M} & {950M}  & {950M} & {950M} & {950M} \\
{Final FLOPs} & {nan} & {949M} & {933M} & {975M} & {920M} & {916M} & {924M} \\
{FLOPs variation} & {624M} & {246M} & {219M} & {168M} & {81M} & {71M} & {30M} \\
{Top-1 Acc } & {nan}  & {73.05} & {73.31}  & {73.50}  & {73.50}  & {72.94} & {72.85}\\
\noalign{\vspace{1.5pt}}
\hline
\hline
\end{tabular}}
\end{table}

\subsection{Ablation Studies}

\subsubsection{Multi-Trial Experiment}
\label{sec:multi_trial}
For measuring the stability of the proposed algorithm, we further conduct a multi-trial experiment. We run the experiment of jointly pruning channel/spatial/depth dimensions on MobileNet V1 targeting at 150M FLOPs three times, and obtain the final accuracy of pruned networks to be 67.64 \%, 67.51\% and 67.69\%, with the variance of 0.075\%. This result reflects that the proposed algorithm is stable and has little variance regarding weight initialization and randomness during training.

\subsubsection{Lipschitz continuity with respect to $\vec{v}$}
\label{sec:lipschitz}
We conduct experiments on MobileNet V1/V2 and ResNet to calculate the minimum positive real constant $K$ satisfying Eq.~\ref{eq:lipschitz}.
We track the maximum value of $ \frac{||\mathcal{E}(\vec{\mu}+\vec{n}) -  \mathcal{E}(\vec{\mu})||}{||\vec{n}||}$, \textit{i.e.}, the lower bound of $K$, during the experiment, which is presented in Table.~\ref{table:lipschitz}. We find that, the lower bound of $K$ peaks at early iterations and decreases with the sigma decreasing. For all three networks in our experiment, the $K$ value is bounded, supporting that the neural network is \textit{Lipschitz} continuous with respect to $\vec{v}$.

\subsubsection{The effect of regularization parameter $\rho$}
\label{sec:rho}
We empirically choose the value of $\rho$ to be 1\textit{e}-8 for the MobileNet-based experiments and 2\textit{e}-9 for the ResNet-based experiments. The reason behind the value choice of $\rho$ is to match the magnitude of the classification loss and FLOPs loss. The initial classification loss for 1000-class classification problem is $ln(1000) \approx 6.9$, and we choose the value of $\rho$ to have $(\rho ||\mathcal{C}(\vec{v})-constraint||)^2 \approx 1\times10^1$. For further investigating the effect of $\rho$, we conducted an ablation study on ResNet-50 targeting at $constrain$ = 1G. As shown in Table~\ref{table:rho}, the final FLOPs loss is regularized to the order of 1\textit{e}-3. When $\rho$ is large, the final FLOPs is very close to the target FLOPs, and the final accuracy is lower. This is because large $\rho$ makes the FLOPs loss dominant the total loss and restricts the search algorithm from exploring the architectures with higher accuracy but slightly bigger deviation to the target FLOPs. We find that in this experiment, choosing $\rho$ to be 2\textit{e}-9 or 1\textit{e}-9 works well.

\subsubsection{The effect of standard deviation $\sigma$}
\label{sec:sigma}
In updating the pruning vector, we sample pruning vectors obeying Gaussian distribution with a standard deviation $\sigma$, \textit{i.e.}, $p_{\mu}(\vec{v})\sim \mathcal{N}(\vec{\mu},\sigma)$. To estimate the gradient precisely, we want the sampled architectures to have sufficient variance while retaining within the local region of the current pruning vector. This is controlled by $\sigma$. Here, we conduct an ablation study, to investigate the effect of $\sigma$ on architectural variation and the final performance. As shown in Table~\ref{table:sigma}, the FLOPs variation grows with the increase in the value of $\sigma$. If the sigma is overly large, the training process becomes unstable and hard to converge, while if the sigma is too small, it fails to explore plentiful enough architectures and thus yields inferior accuracy. We find that setting the initial value of $\sigma$ to be 0.01$\sim$0.015 yields desirable final accuracy.

\section{Conclusion}
In this work, we focused on the joint multi-dimension pruning which naturally has broader pruning space on spatial, depth and channel for digging out better configurations than the isolated single dimension solution. We have proposed to use stochastic gradient estimation to optimize this problem and further introduced a weight sharing strategy to avoid repeatedly training multiple models from scratch. Our results on large-scale ImageNet dataset with MobileNet V1\&V2\&V3 and ResNet outperformed previous state-of-the-art pruning methods with significant margins.

\section*{Acknowledgement}

The authors would like to acknowledge the National Key Research and Development Program of China (No. 2020AAA0105200), Beijing Academy of Artificial Intelligence (BAAI) and the sponsorship of Research Grants Council of Hong Kong. This work was partially supported by Hong Kong General Research Fund (GRF) 16203319 and an RGC research grant under the National Nature Science Foundation of China/Research Grants Council Joint Research Scheme and AI Chip Center for Emerging Smart Systems (ACCESS), Hong Kong SAR.

\bibliographystyle{plain}
\bibliography{main}

\end{document}